\documentclass[letterpaper,10pt,conference]{ieeeconf}
\IEEEoverridecommandlockouts
\usepackage[T1]{fontenc}
\usepackage{amsmath,amssymb,booktabs,graphicx,url,xcolor}
\usepackage{balance}
\usepackage{tikz}
\usepackage{xcolor}
\usepackage{graphicx}
\usepackage{algorithm}
\usepackage{algpseudocode}
\usetikzlibrary{arrows.meta,positioning,calc,fit}

\definecolor{myblue}{RGB}{59,130,246}
\definecolor{mygreen}{RGB}{34,197,94}
\definecolor{myorange}{RGB}{249,115,22}
\definecolor{mypurple}{RGB}{139,92,246}
\definecolor{mygray}{RGB}{71,85,105}

\definecolor{lightblue}{RGB}{239,246,255}
\definecolor{lightgreen}{RGB}{240,253,244}
\definecolor{lightorange}{RGB}{255,247,237}
\definecolor{lightpurple}{RGB}{245,243,255}
\definecolor{lightgray}{RGB}{248,250,252}

\usepackage{makecell}
\usepackage{amsmath}
\usepackage{svg}

\title{Learning to See through Action: Predictive Curiosity for Robots}
\author{Stelios Zarifis\textsuperscript{1,2,3}, Panagiotis Filntisis\textsuperscript{1,2,3}, George Retsinas\textsuperscript{2}, and Petros Maragos\textsuperscript{1,2,3}\\[0.5em]
	\textsuperscript{1}HERON - Hellenic Robotics Center of Excellence, Athena Research Center, Greece \\
	\textsuperscript{2}Institute of Robotics, Athena Research Center, Athens, Greece \\
	\textsuperscript{3}National Technical University of Athens, School of Electrical \& Computer Engineering, Greece \\ [0.5em]
	\texttt{\{s.zarifis, pfilntisis, george.retsinas, petros.maragos\}@athenarc.gr}
}

\begin{document}
\bstctlcite{bstctl:nodash}
\maketitle

\begin{abstract}
    Humans and animals do not learn vision from a fixed set of data: they move, interact, and learn from the visual consequences of their actions. Motivated by this developmental view, we study whether robots can improve visual learning by actively collecting experience through curiosity-driven interaction. We introduce DySCOVER, a self-supervised framework for active robot visual learning. A robot observes RGB frames, learns a visual autoencoder that encodes and reconstructs them, and trains a recurrent world model to predict the next visual latent state conditioned on the current state and action. The agent's policy is trained only with intrinsic rewards: reconstruction error encourages visually novel observations, while the world model's error encourages transitions whose consequences are not yet predictable. This prediction error biases the policy toward more intricate trajectories that enrich the agent's set of experiences with observations beyond those encouraged by visual novelty alone. Our experiments conducted with robotic arms as the embodied agents show that adding predictive curiosity improves the vision model over using visual curiosity alone. Across variations in initial conditions, experience capacity, embodiment, and scene content, adding predictive curiosity consistently improved visual learning. These results show that predictive curiosity can improve active visual learning beyond reconstruction-based curiosity alone.
\end{abstract}

\section{Introduction}

Humans and animals do not learn vision from a fixed dataset of images. From the earliest stages of development, infants move their eyes, hands, and bodies to interact with their environment and learn from the consequences of their actions. This loop of observing, acting, predicting, and updating internal models motivates open-ended learning in robotics~\cite{schmidhuber1991curious,schmidhuber2006developmental,oudeyer2007intrinsic}. An embodied agent should therefore not only learn from observations, but also \textit{perform actions that generate experience useful for improving its internal models}.

Curiosity-driven learning guides exploration through intrinsic signals such as state novelty, prediction error, model disagreement, information gain, and learning progress~\cite{burda2019,pathak2017curiosity,pathak2019disagreement,houthooft2016vime,kim2020active}. These signals have also supported visual representation learning from self-collected experience, as in Curious Representation Learning~\cite{du2021curious}. 

Recently, Push, See, Predict~\cite{konstantaropoulos2025push} showed that
prediction error can be used to collect informative object interactions and
improve visual reconstruction. However, it studies a staged object-centric
setting built around explicit object decomposition and predefined pushing
actions. In contrast, we study a continually updated learning loop that
operates directly on learned visual latents, without requiring object segmentation, in which the policy, visual model, and world model co-adapt from the agent's own interactions.

Specifically, we investigate the question: \textit{\textbf{does predictive curiosity improve visual learning beyond what reconstruction-based curiosity alone can achieve?}} This visual curiosity rewards observations that are difficult to reconstruct. Predictive curiosity instead rewards transitions whose consequences are difficult to predict. We argue that the predictive signal can bias the policy towards actions that cause more intricate motions, leading to useful experiences for the agent, that a pure vision-novelty reward may not achieve. This motivates us to study whether prediction error can provide a complementary exploration signal for collecting visual experience that improves the agent's visual model.

To study this question, we introduce \textbf{DySCOVER} (DynamicS-informed CuriOsity for Visual Exploration and Representation-learning), a self-supervised framework for active visual representation learning in embodied agents. DySCOVER couples three components that are updated throughout exploration: a visual autoencoder that learns from the collected image observations, a recurrent world model that predicts action-conditioned transitions in the visual latent space, and a recurrent policy that decides what experience to collect. The policy is trained only with intrinsic reward: reconstruction error encourages visually novel observations, while prediction error encourages transitions whose consequences are difficult to predict. The prediction error of the world model influences the policy and therefore the visual experience on which the vision model is trained. Our extensive experimental results, across various settings, consistently show that adding predictive curiosity improves visual reconstruction.
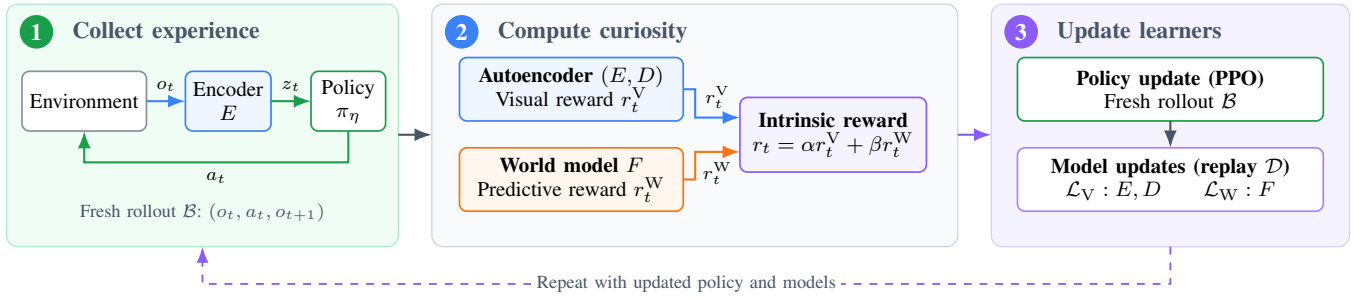
\begin{figure*}[t]
	\centering
	\resizebox{\textwidth}{!}{%
		\begin{tikzpicture}[
			x=1cm,
			y=1cm,
			font=\footnotesize,
			>=Latex,
			box/.style={
				rounded corners=1mm,
				line width=0.65pt,
				draw=mygray!65,
				fill=white,
				align=center,
				inner sep=3pt,
				outer sep=0pt
			},
			flow/.style={
				-Latex,
				line width=0.8pt,
				draw=mygray
			},
			act/.style={flow, draw=mygreen!80!black},
			vis/.style={flow, draw=myblue},
			pred/.style={flow, draw=myorange},
			rew/.style={flow, draw=mypurple},
			stagebadge/.style={
				circle,
				minimum size=4.5mm,
				inner sep=0pt,
				text=white,
				font=\bfseries\small
			},
			stagehead/.style={
				anchor=west,
				font=\bfseries\small,
				text=mygray
			}
			]
			
			
			\filldraw[
			rounded corners=1.5mm,
			fill=lightgreen,
			draw=mygreen!35,
			line width=0.5pt
			] (0,0.35) rectangle (5.25,3.60);
			
			\filldraw[
			rounded corners=1.5mm,
			fill=lightgray,
			draw=mygray!25,
			line width=0.5pt
			] (5.70,0.35) rectangle (12.75,3.60);
			
			\filldraw[
			rounded corners=1.5mm,
			fill=lightpurple,
			draw=mypurple!30,
			line width=0.5pt
			] (13.20,0.35) rectangle (18.00,3.60);
			
			\node[stagebadge, fill=mygreen!80!black]
			at (0.40,3.23) {1};
			\node[stagehead]
			at (0.76,3.23) {Collect experience};
			
			\node[stagebadge, fill=myblue]
			at (6.10,3.23) {2};
			\node[stagehead]
			at (6.46,3.23) {Compute curiosity};
			
			\node[stagebadge, fill=mypurple]
			at (13.60,3.23) {3};
			\node[stagehead]
			at (13.96,3.23) {Update learners};
			
			
			\node[
			box,
			minimum width=16mm,
			minimum height=8mm
			] (env) at (1.04,2.30)
			{Environment};
			
			\node[
			box,
			draw=myblue,
			fill=lightblue,
			minimum width=11mm,
			minimum height=8mm
			] (enc) at (2.97,2.30)
			{Encoder\\$E$};
			
			\node[
			box,
			draw=mygreen!80!black,
			minimum width=10mm,
			minimum height=8mm
			] (pol) at (4.58,2.30)
			{Policy\\$\pi_\eta$};
			
			\draw[vis]
			(env.east)
			-- node[above, font=\scriptsize] {$o_t$}
			(enc.west);
			
			\draw[act]
			(enc.east)
			-- node[above, font=\scriptsize] {$z_t$}
			(pol.west);
			
			\draw[act]
			(pol.south)
			-- (4.58,1.48)
			-- node[below, font=\scriptsize] {$a_t$}
			(1.04,1.48)
			-- (env.south);
			
			\node[
			font=\scriptsize,
			text=mygray,
			align=center
			] at (2.625,0.81)
			{Fresh rollout $\mathcal B$: $(o_t,a_t,o_{t+1})$};
			
			
			\node[
			box,
			draw=myblue,
			fill=lightblue,
			minimum width=30mm,
			minimum height=8.5mm
			] (rv) at (7.57,2.46)
			{\textbf{Autoencoder} $(E,D)$\\
				Visual reward $r_t^{\mathrm V}$};
			
			\node[
			box,
			draw=myorange,
			fill=lightorange,
			minimum width=30mm,
			minimum height=8.5mm
			] (rw) at (7.57,1.25)
			{\textbf{World model} $F$\\
				Predictive reward $r_t^{\mathrm W}$};
			
			\node[
			box,
			draw=mypurple,
			fill=lightpurple,
			minimum width=25mm,
			minimum height=10mm
			] (reward) at (11.08,1.85)
			{\textbf{Intrinsic reward}\\
				$r_t=\alpha r_t^{\mathrm V}+\beta r_t^{\mathrm W}$};
			
			\draw[vis]
			(rv.east)
			-- (9.23,2.46)
			|- node[pos=0.75, above, font=\scriptsize]
			{$r_t^{\mathrm V}$}
			([yshift=2.3mm]reward.west);
			
			\draw[pred]
			(rw.east)
			-- (9.23,1.25)
			|- node[pos=0.75, below, font=\scriptsize]
			{$r_t^{\mathrm W}$}
			([yshift=-2.3mm]reward.west);
			
			\draw[flow] (5.25,1.85) -- (5.70,1.85);
			\draw[rew] (12.75,1.85) -- (13.20,1.85);
			
			
			\node[
			box,
			draw=mygreen!80!black,
			minimum width=41mm,
			minimum height=8.5mm
			] (ppo) at (15.60,2.46)
			{\textbf{Policy update (PPO)}\\
				Fresh rollout $\mathcal B$};
			
			\node[
			box,
			draw=mypurple!75,
			minimum width=41mm,
			minimum height=8.5mm
			] (models) at (15.60,1.25)
			{\textbf{Model updates (replay $\mathcal D$)}\\
				$\mathcal L_{\mathrm V}: E,D
				\qquad
				\mathcal L_{\mathrm W}: F$};
			
			\draw[flow] (ppo.south) -- (models.north);
			
			
			\draw[
			-Latex,
			dashed,
			draw=mypurple,
			line width=0.7pt
			]
			(15.60,0.35)
			-- (15.60,-0.15)
			-- node[
			midway,
			fill=white,
			inner sep=2pt,
			font=\scriptsize,
			text=mygray
			] {Repeat with updated policy and models}
			(2.625,-0.15)
			-- (2.625,0.35);
			
		\end{tikzpicture}%
	}
    \vspace{-0.7cm}
	\caption{
		DySCOVER Overview.
		(1) The recurrent policy collects
		action--observation transitions using the visual representation as state.
		(2) Intrinsic curiosity is calculated using the errors of the vision and world models.
		(3) Policy is updated online, visual and world models are updated using replayed experience.
	}
	\label{fig:method}
    \vspace{-1em}
\end{figure*}

\section{Related Work}

\subsection{Developmental Learning and Artificial Curiosity}

The idea that an agent should use intrinsic motivation to choose experiences useful for learning has been studied extensively. Early curiosity-driven systems coupled predictive models with controllers that sought improvements in their predictions~\cite{schmidhuber1991curious,schmidhuber2006developmental}. This idea has been extended through learning and competence progress, directing exploration toward experiences from which the agent can learn rather than simply those with the largest error~\cite{oudeyer2007intrinsic,baranes2013}. The Random Network Distillation (RND) method rewards errors in predicting the output of a fixed random network, providing an observation-based novelty signal~\cite{burda2019}. The Intrinsic Curiosity Module (ICM) uses prediction error as curiosity: the agent is rewarded when its actions lead to outcomes that are difficult to predict and possible to be predicted, in a learned feature space~\cite{pathak2017curiosity}. Other works refine this idea by rewarding learning progress rather than raw prediction error. Active World Model Learning, for example, compares the current prediction loss with that of a slowly updated reference model, so that the agent prefers situations where its model is improving~\cite{kim2020active}. Curiosity-Critic follows a related motivation by learning a baseline for expected prediction error and rewarding deviations from this baseline, reducing the tendency to chase errors that remain high but not informative~\cite{bhaskara2026}.

\subsection{Visual Learning from Self-Generated Experience}
Physical interaction can provide supervision for visual learning. The Curious Robot learns visual representations from robotic pushing, poking, grasping, and object observations~\cite{pinto2016curious}. Curious Representation Learning (CRL) explicitly couples representation learning and exploration: a policy maximizes the visual learner's contrastive loss, while the learner minimizes that loss on the collected observations~\cite{du2021curious}. Haber et al. jointly train a self- and world-model to perform on external tasks~\cite{haber2018learning}.

Push, See, Predict is close to our approach~\cite{konstantaropoulos2025push}, since it also uses prediction error to collect informative visual experience. However, it studies a staged, object-centric pipeline in which Slot Attention decomposes each frame into a predefined number of object representations and segmentation masks, which are subsequently used by a graph-based world model and the exploration policy. In contrast, DySCOVER operates directly on a learned latent visual space. Moreover, we study whether dynamics-aware curiosity provides an advantage over an already active visual-only curiosity policy, within a continually updated loop where the visual and world model, and exploration policy co-adapt throughout exploration, rather than primarily over random or heuristic data collection.

\subsection{World Models for Exploration and Control}
World models also support exploration and control beyond visual data acquisition. VIME rewards information gain about learned dynamics~\cite{houthooft2016vime}, while disagreement-based exploration uses variation among predictive models~\cite{pathak2019disagreement}. Plan2Explore trains an exploration policy within a learned world model, using an ensemble of one-step predictors to estimate latent disagreement~\cite{sekar2020plan2explore}. PlaNet and Dreamer use latent dynamics for planning and policy learning from imagined trajectories~\cite{hafner2019planet,hafner2025dreamerv3}. BYOL-Explore jointly learns visual representations, dynamics, and exploration through a bootstrapped latent prediction objective~\cite{guo2022}. In DySCOVER, the world model offers a complementary curiosity signal to guide the policy towards collecting informative experiences to improve its vision model. We evaluate the effectiveness of our method primarily through the reconstruction quality of the vision model on an independently created, unseen evaluation set of visual observations.

\section{Methodology}
\label{sec:methodology}

\subsection{Objective and Learning Setting}

We study active visual representation learning for an embodied agent. At time $t$, the agent receives a visual frame observation $o_t\in\mathcal{O}$, selects a continuous action $a_t\in\mathcal{A}$, and observes the next visual frame $o_{t+1}$. The policy state is the latent visual representation learned from the visual observations; no other simulator variables are provided to the policy or used for the exploration. The objective of the agent is to learn how to interact with the environment in order to collect data that improves its visual model.

This objective is viewed as a closed loop. The policy selects actions based on the agent's current internal models of the world, and these actions determine which observations and transitions are collected. The collected experience is stored in a replay buffer and used to update the agent's internal models and policy. As these models improve, both the state representation and the regions where the models are uncertain or incomplete change so the policy must adapt towards collecting the data distribution that is most informative for learning. An overview of this closed-loop learning process is shown in Figure~\ref{fig:method}.


We begin with self-body exploration: the scene contains no objects, so visual change is attributed to the robot’s own motion. This controlled setting lets the agent learn the relationship between its actions and the resulting changes in the camera view, without mixing them with contact or object motion. When objects are introduced later, the model already contains expectations about its body motions, allowing predictive curiosity to focus more on changes caused by interaction with the environment. It follows a simple developmental intuition: the agent first learns the effects of its own movements, then learns what happens when acting on objects around it.

\subsection{Visual Representation and Curiosity}
Let $E_\theta$ and $D_\theta$ denote an image encoder and decoder, where $\theta=(\theta_E,\theta_D)$ are the parameters of the full visual autoencoder. The visual representation and reconstruction of an observation $o_t$ are
\begin{equation}
	z_t=E_{\theta_E}(o_t), \qquad \hat o_t=D_{\theta_D}(z_t).
\end{equation}
The representation is learned in a self-supervised way, using a combination of pixel and perceptual reconstruction terms~\cite{johnson2016perceptual}:
\begin{equation}
	\mathcal{L}_{\mathrm{V}}(o_t;\theta)=
	\lambda_{\mathrm{pix}}\,\ell_{\mathrm{pix}}(o_t,\hat o_t)
	+\lambda_{\mathrm{per}}\,\ell_{\mathrm{per}}(o_t,\hat o_t).
	\label{eq:ae-loss}
\end{equation}
The reconstruction loss, evaluated on the obtained observation after the agent applies an action, defines the visual error:
\begin{equation}
	e^{\mathrm{V}}_t = \mathcal{L}_{\mathrm{V}}(o_{t+1};\theta).
	\label{eq:visual-error}
\end{equation}
For visual-only curiosity, this error is used directly as the intrinsic reward:
\begin{equation}
	r^{\mathrm{V}}_t = e^{\mathrm{V}}_t.
	\label{eq:visual-reward}
\end{equation}
The visual-only baseline uses Eq.~\eqref{eq:visual-reward} as its complete reward. In this case, exploration is driven solely by frame novelty.

\subsection{Action-Conditioned World Model}

Our hypothesis is that using a world model (WM), which predicts the outcomes of actions, alongside the visual curiosity, can enable the agent to discover more intricate motions, leading to new states that visual curiosity alone cannot provide the proper reward signal to reach. We use a recurrent predictor with parameters $\phi=(\phi_F,\phi_G)$, which receives the current visual representation as the latent state $z_t$, the action, and a hidden state $h_t$ that encodes history (since past state-action trajectories compile momentum on the robot's joints, which cannot be captured by the visual representation of a single frame):

\begin{equation}
	\tilde z_{t+1}=F_\phi(z_t,a_t,h_t),
	\qquad h_{t+1}=G_\phi(z_t,a_t,h_t).
	\label{eq:wm}
\end{equation}
The WM is trained to minimize two targets. The primary target aims to match the predicted latent $\tilde{z}_{t+1}$ with the actual next observed latent $z_{t+1}$. The auxiliary target matches the decoded prediction with the decoded target latent rather than the raw next observation, so that the resulting prediction error is not biased by reconstruction errors of the vision model -- a property especially useful for reward shaping explained below:
\begin{equation}
	\mathcal{L}_{\mathrm{W}} =
	\ell_z\!\left(z_{t+1},\tilde z_{t+1}\right)
	+\lambda_{\mathrm{dec}}\,
	\ell_{\mathrm{pix}}\!\left(D_{\theta_D}(z_{t+1}),D_{\theta_D}(\tilde z_{t+1})\right).
	\label{eq:wm-loss}
\end{equation}
During optimization, the world model learns to predict the visual representation, but its loss does not flow into the vision model. The vision model is updated separately using Eq.~\eqref{eq:ae-loss}.

\begin{algorithm}[t]
    	\caption{DySCOVER: Active Visual Learning}
	\label{alg:egomotion}
	\small
	\begin{algorithmic}[1]
		\Require Visual encoder--decoder $(E_{\theta_E},D_{\theta_D})$;
		recurrent world model $(F_\phi,G_\phi)$;
		recurrent actor--critic $\pi_\eta$
		\Require $N$ parallel environments, rollout length $T$,
		reward weights $\alpha,\beta$
		
		\State Initialize models, environments, and replay buffer $\mathcal{D}$
		
		\While{interaction budget remains}
		\State $\mathcal{B} \gets \emptyset$
		\Comment{Fresh on-policy rollout}
		
		\State \textbf{Collect experience with fixed model parameters}
		\For{$t=0,\ldots,T-1$}
		\Statex \hspace{\algorithmicindent}
		\emph{In parallel across the $N$ environments:}
		\State $z_t \gets E_{\theta_E}(o_t)$
		\State Sample $a_t \sim
		\pi_\eta(\cdot \mid z_t,h_t^\pi)$
		\State $\tilde z_{t+1} \gets
		F_{\phi_F}(z_t,a_t,h_t^{\mathrm{W}})$
		\State Execute $a_t$ and observe $o_{t+1}$
		\State $z_{t+1} \gets E_{\theta_E}(o_{t+1})$; \;\; $e_t^{\mathrm{V}} \gets \mathcal{L}_{\mathrm{V}}(o_{t+1};\theta)$; \;\; $r_t^{\mathrm{V}} \gets e_t^{\mathrm{V}}$
		\State $e_t^{\mathrm{W}} \gets
		\mathcal{L}_{\mathrm{W}}(z_t,a_t,z_{t+1};\phi)$; \quad $r_t \gets \alpha r_t^{\mathrm{V}} + \beta r_t^{\mathrm{W}}$
		\State \textit{Computed $r_t^{\mathrm{W}}$ from $e_t^{\mathrm{W}}$ using Eq.~\ref{eq:wm-reward} or Eq.~\ref{eq:floor-reward}}
		\State Store transition, reward, and PPO
		rollout statistics in $\mathcal{B}$
		\State Store $(o_t,a_t,h_t^W,o_{t+1})$ in $\mathcal{D}$
		\State Advance observations and recurrent states;
		\State reset at episode boundaries
		\EndFor
		
		\State \textbf{Update the collection policy}
		\State \quad Compute advantages and returns from $\mathcal{B}$
		\State \quad Update $\eta$ with recurrent PPO on $\mathcal{B}$,
		keeping $\theta$ fixed
		
		\State \textbf{Update the visual and predictive models}
		\State \quad Update $\theta$ on replayed images using
		$\mathcal{L}_{\mathrm{V}}$
		in Eq.~\eqref{eq:ae-loss}
		\State \quad Re-encode replay sequences using
		the updated $E_{\theta_E}$
		\State \quad Update $\phi$,
		keeping $E_{\theta_E}$ and $D_{\theta_D}$ fixed
		\EndWhile
		
	\end{algorithmic}
\end{algorithm}

\subsection{Dynamics-Aware Intrinsic Reward}

The raw WM reward is defined analogously: the WM's prediction error is used directly as an intrinsic reward,
\begin{equation}
	r^{\mathrm{W}}_t = e^{\mathrm{W}}_t,
	\qquad
	e^{\mathrm{W}}_t = \mathcal{L}_{\mathrm{W}}(z_t,a_t,z_{t+1};\phi).
	\label{eq:wm-reward}
\end{equation}

DySCOVER combines frame-wise visual curiosity with action-conditioned predictive curiosity:
\begin{equation}
	r_t = \alpha\,r^{\mathrm{V}}_t
	+\beta\,r^{\mathrm{W}}_t,
	\qquad \alpha,\beta\geq 0.
	\label{eq:combined-reward}
\end{equation}
where the weights $\alpha, \beta$ are chosen to balance the raw reward terms $r^{\mathrm{V}}_t, r^{\mathrm{W}}_t$ which depend on the environment and the model architectures used.

In order to reduce the policy bias toward high-magnitude actions caused by the world-model loss term (as high magnitude actions tend to cause large transitions in the latent space, therefore larger prediction errors), we use an adjusted version of the WM reward. We introduce a slowly updated predictor $b_\psi$, with parameters $\psi$, that estimates a conservative reference level for the expected WM prediction error under the current latent state and action. In our implementation, $b_\psi$ is a small predictor with parameters $\psi$, trained by regression to predict the observed WM error from $x_t=(z_t,a_t,h_t)$ using lagged stored transitions (meaning from previous update cycles). Its output is constrained to be positive by construction, and the logarithm uses a small positive $\epsilon$ for numerical stability:
\begin{equation}
	\begin{aligned}
		\bar r^{\mathrm{W}}_t
		&=
		\max\left(0, \, \log(e^{\mathrm{W}}_t+\epsilon) - \log(b_\psi(x_t)+\epsilon)\right), \\
		&\phantom{={}} x_t = (z_t, a_t, h_t).
	\end{aligned}
	\label{eq:floor-reward}
\end{equation}
This is a log-ratio between the current prediction error and the predicted reference error. The ratio makes the reward depend on the prediction error relative to its expected level, while the logarithm compresses large ratios, keeping the reward on a smoother scale. Therefore, the reward becomes large when the current transition is unexpectedly difficult for the WM to predict, but remains small when the error is consistently high for that context. When using the adjusted WM reward, we replace $r_t^{\mathrm W}$ in Eq.~\eqref{eq:combined-reward} with $\bar r_t^{\mathrm W}$ from Eq.~\eqref{eq:floor-reward}.

\subsection{Policy optimization and alternating updates}

The policy is a recurrent actor--critic $\pi_\eta(a_t \mid z_t,h_t^\pi)$, with parameters $\eta$, and is optimized using Proximal Policy Optimization (PPO)~\cite{schulman2017ppo}. Both the actor and critic are recurrent to exploit temporal context through the hidden state $h_t^\pi$, which again cannot be encoded in the visual representation.




The learning occurs in cycles. First, the current policy is used to collect observation-action rollouts. Second, the policy is updated using the observed rewards. Third, the vision and world models are updated using the collected data in the replay buffer. The updated models are used for the next cycle. The procedure is summarized in Algorithm~\ref{alg:egomotion}.

\subsection{Model Architectures}

The visual model is implemented as a spatial-bottleneck autoencoder. It uses a ResNet encoder without pretraining and retains only the deepest encoder feature map as the visual embedding. The decoder receives only this bottleneck and reconstructs the image. We intentionally do not use encoder--decoder skip connections, so high-level information does not bypass the bottleneck, since the latent state used by the policy and world model should carry the full visual information.

The world model is implemented as a convolutional recurrent model that predicts the next visual latent state from the current latent state and executed action. The visual representation retains its spatial structure, while the action is projected through a learned action embedding over the same spatial grid. The concatenated tensor is processed by a convolutional GRU (ConvGRU~\cite{ballas2016convgru}), producing a prediction for the next latent and a hidden state that encodes the temporal information. The collection policy is implemented as a recurrent actor--critic model.

The visual and world model are trained using the losses defined in Eq.~\eqref{eq:ae-loss} and Eq.~\eqref{eq:wm-loss}, respectively.

\section{Experimental Results}
\label{sec:results}

\begin{figure}[t]
	\centering
	\textbf{Environments used in the study}\\[2mm]
	
	\begin{minipage}[c]{0.24\linewidth}
		\centering
		\scriptsize UR10e with\\Shadow Hand\\[1mm]
		\includegraphics[width=\linewidth]{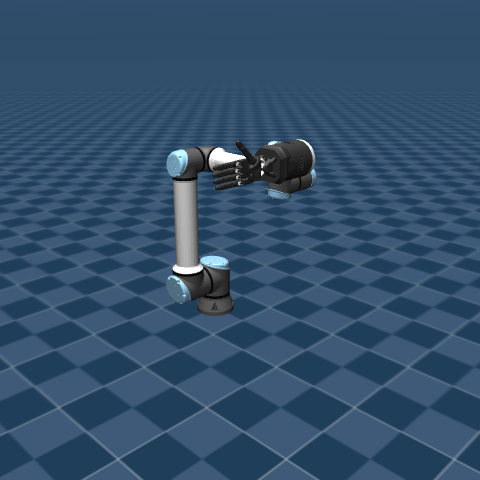}
	\end{minipage}
	\hfill
	\begin{minipage}[c]{0.24\linewidth}
		\centering
		\scriptsize Franka with\\Shadow Hand\\[1mm]
		\includegraphics[width=\linewidth]{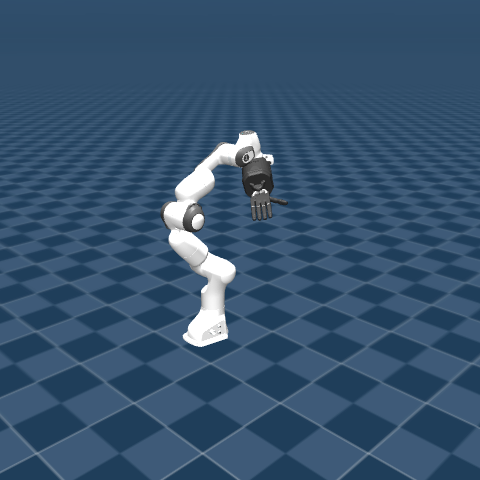}
	\end{minipage}
	\hfill
	\begin{minipage}[c]{0.24\linewidth}
		\centering
		\scriptsize UR10e with\\Gripper\\[1mm]
		\includegraphics[width=\linewidth]{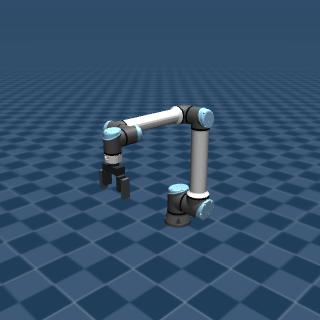}
	\end{minipage}
	\hfill
	\begin{minipage}[c]{0.24\linewidth}
		\centering
		\scriptsize UR10e with\\Objects\\[1mm]
		\includegraphics[width=\linewidth]{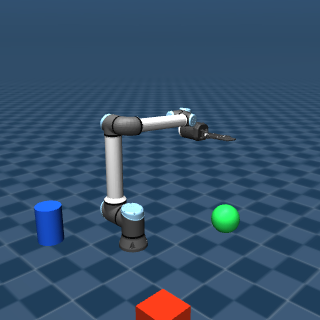}
	\end{minipage}
	
	\caption{Robot environments used in this study: UR10e arm with an anthropomorphic Shadow Hand (first), Franka arm (second), UR10e arm with a different gripper (third), and an object-augmented scene (fourth) for testing adaptability on new environments.}
	\label{fig:robots}
\end{figure}

\subsection{Experimental Setup}
We run all experiments on a MuJoCo simulation environment which includes the embodied agent and a static RGB camera for the visual observations, which are rendered at $320\times320$ pixels. We run 10 environments in parallel, and each collection cycle contains 400 steps per environment, so in total 4,000 new transitions per cycle. The replay buffer holds 200,000 frames in the main case-study and 16,000 frames in the compact-buffer case-study. Within each ablation, the simulator, camera, random initialization reset, model architectures, replay capacity, update schedule, and evaluation data are kept fixed.
    
We compare five exploration strategies. \textit{Random exploration} samples actions without a learned policy, serving as a reference for unguided data collection. The \textit{visual-only curiosity} baseline uses the reconstruction error of the visual model as the intrinsic reward. \textit{WM-only} uses only the world-model prediction error as reward; although this signal is predictive, it is still implicitly visual because the world model operates on visual latents. \textit{DySCOVER with raw WM reward} augments visual curiosity with the raw prediction error of the world model. Finally, \textit{DySCOVER with adjusted WM reward} replaces the raw prediction term with the baseline-adjusted reward of Eq.~\eqref{eq:floor-reward}, which measures the prediction error relative to a learned reference level. Unless otherwise stated, DySCOVER refers to the adjusted-WM variant.

\begin{figure}[t]
    \centering
    \includesvg[width=\columnwidth]{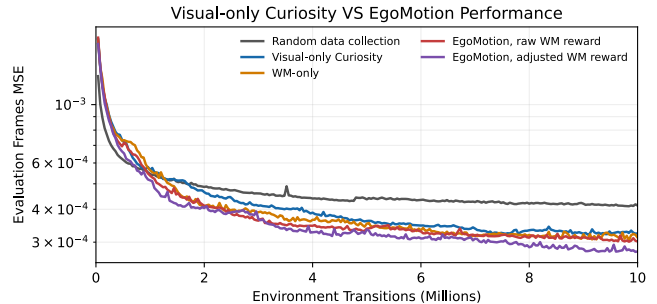}
    \caption{Evaluation frame MSE during 10 million training steps for the UR10e with ``Shadow Hand''. Using the WM reward term, the convergence is faster and to lower reconstruction error than visual-only curiosity.}
    \label{fig:primary-mse}
\end{figure}

\begin{table}[t]
	\centering
	\caption{DySCOVER performance}
	\label{tab:reconstruction-full-buffer}
	\small
	\resizebox{\columnwidth}{!}{%
        \begin{tabular}{cccc}
			\toprule
			Method & \makecell{MSE $\downarrow$} & \makecell{PSNR (dB) $\uparrow$} & \makecell{MSE reduction \\ w.r.t. baseline} $\uparrow$ \\
			\midrule
			Random data collection & $4.08\times10^{-4}$ & $33.90$ & $-25.9\%$ \\
			Visual-only curiosity (baseline) & $3.24\times10^{-4}$ & $34.89$ & --- \\
			World Model reward only & $3.12\times10^{-4}$ & $35.06$ & $3.7\%$ \\
			DySCOVER, raw WM reward & $3.02\times10^{-4}$ & $35.20$ & $6.8\%$ \\
			\textbf{DySCOVER, adjusted WM reward} & $\mathbf{2.77\times10^{-4}}$ & $\mathbf{35.58}$ & $\mathbf{14.5\%}$ \\
			\bottomrule
	\end{tabular}}
\end{table}

\subsection{Evaluation Measures}

Our primary evaluation measures visual reconstruction on a set of RGB frames generated independently of the RL training loop. The set contains 3,000 images rendered from random robot states from the training environment distribution. These frames are generated once and reused across methods, so the metrics compare the visual models on the same observations. We evaluate using mean squared error (MSE) and peak signal-to-noise ratio (PSNR). We also record policy and rollout statistics during training to examine qualitatively how the world-model 
influences exploration.

\subsection{Primary Exploration Comparison}

\begin{figure}[t]
	\centering
	\begin{minipage}[b]{0.3\linewidth}
		\centering
		\footnotesize\strut Ground Truth \\
		\includegraphics[width=\linewidth]{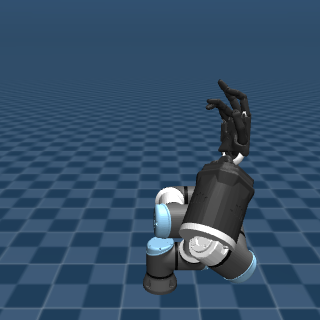}
	\end{minipage}
	\hfill
	\begin{minipage}[b]{0.3\linewidth}
		\centering
		\footnotesize\strut Visual-only Curiosity \\
		\includegraphics[width=\linewidth]{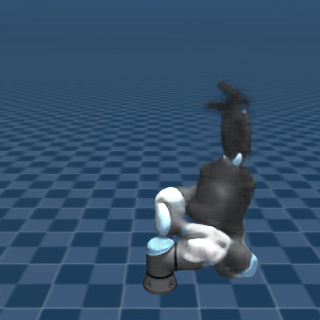}
	\end{minipage}
	\hfill
	\begin{minipage}[b]{0.3\linewidth}
		\centering
		\footnotesize\strut DySCOVER \\
		\includegraphics[width=\linewidth]{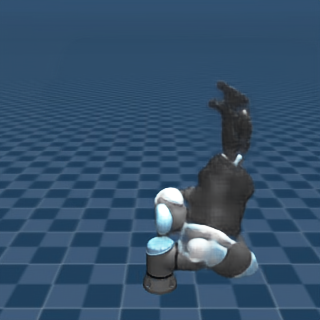}
	\end{minipage}
	
	\vspace{1mm}
	
	\begin{minipage}[b]{0.3\linewidth}
		\hfill
		\includesvg[height=\linewidth, keepaspectratio]{images/rank_06_sample_0146/reconstruction_error_colorbar.svg}
	\end{minipage}
	\hfill
	\begin{minipage}[b]{0.3\linewidth}
		\centering
		\footnotesize\strut Visual-only Heatmap \\
		\includesvg[width=\linewidth]{images/rank_06_sample_0146/visual_only_curiosity_squared_error_heatmap.svg}
	\end{minipage}
	\hfill
	\begin{minipage}[b]{0.3\linewidth}
		\centering
		\footnotesize\strut DySCOVER Heatmap \\
		\includesvg[width=\linewidth]{images/rank_06_sample_0146/dyscover_squared_error_heatmap.svg}
	\end{minipage}
	
	\caption{Example reconstruction and error heatmap comparisons on a frame during training.}
    \vspace{-1.8em}
	\label{fig:primary_reconstruction}
\end{figure}

\noindent\textbf{Vision Model Improvement.} Figure~\ref{fig:primary-mse} shows the visual-learning curves of the five exploration strategies over 10 million simulation steps, with the final reconstruction performance shown in Table~\ref{tab:reconstruction-full-buffer}. Figure~\ref{fig:primary_reconstruction} further shows a qualitative reconstruction example, along with heatmaps to highlight the pixel-wise differences.


The visual-only curiosity strategy serves as the main baseline, since it uses only frame reconstruction novelty exploration. Using only the WM prediction error as the intrinsic reward achieves a $3.7\%$ lower reconstruction error than this baseline, showing that predictive curiosity alone can result in useful data collection (the vision model is still trained from the collected images). Adding the raw WM reward on top of visual curiosity gives a larger improvement of $6.8\%$, indicating that prediction error can complement the reconstruction novelty to bias the policy towards collecting useful experiences. The strongest result is obtained with the adjusted WM reward: DySCOVER reduces reconstruction error by $14.5\%$ relative to visual-only curiosity. Finally, random exploration performs $25.9\%$ worse than visual-only curiosity and $47.3\%$ worse than DySCOVER, confirming that curiosity-driven exploration offers significant advantage in visual representation learning.

\noindent\textbf{Convergence Speed.} The difference is also reflected in convergence speed. The reconstruction error reached by visual-only curiosity after 10 million steps is reached by DySCOVER with raw WM reward after approximately 5 million steps, and by DySCOVER with adjusted WM reward after approximately 4 million steps.

\noindent\textbf{Multi-seed experiments.} We also repeated the main experiments with 3 different random seeds to account for stochasticity. The standard deviation of the MSE reconstruction was measured $1.76\%$ of the mean for visual-only curiosity and $4.11\%$ for DySCOVER, indicating result consistency.

\subsection{Exploration with Compact Sized Buffer}

\begin{figure}[t]
	\centering
	\includesvg[width=\columnwidth]{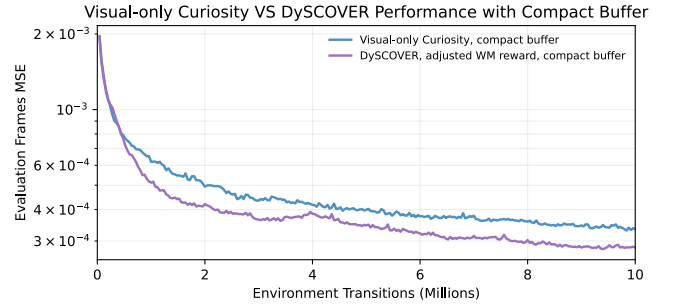}
    \caption{Evaluation frame reconstruction MSE during 10 million training steps with a compact replay buffer. Lower is better. DySCOVER converges faster and reaches lower reconstruction error than visual-only curiosity despite the reduced replay buffer capacity.}
	\label{fig:small-mse}
\end{figure}

We also test the performance of DySCOVER in a compact-buffer setting, shown in Figure~\ref{fig:small-mse}. The replay buffer is reduced from 200k frames to 16k frames, forcing both the visual-only curiosity baseline and DySCOVER to learn from a substantially smaller experience collection. In this lower budget setting, we test whether dynamics-aware curiosity remains useful when the visual model has to update on a limited replay distribution. As seen in Table ~\ref{tab:reconstruction-small-buffer}, at 10M simulation steps, DySCOVER gives a MSE reduction of $15.3\%$, relative to visual-only curiosity, showing that the benefit of the dynamics-aware reward persists even when replay data are highly constrained.

\begin{table}[t]
	\centering
	\caption{Compact-buffer reconstruction performance}
	\label{tab:reconstruction-small-buffer}
	\small
	\resizebox{\columnwidth}{!}{%
		\begin{tabular}{cccc}
			\toprule
			Method & \makecell{MSE $\downarrow$} & \makecell{PSNR (dB) $\uparrow$} & \makecell{MSE reduction \\ w.r.t. baseline} $\uparrow$ \\
			\midrule
			Visual-only curiosity & $3.34\times10^{-4}$ & $34.76$ & --- \\
			\textbf{DySCOVER} & $\mathbf{2.83\times10^{-4}}$ & $\mathbf{35.49}$ & $\mathbf{15.3\%}$ \\
			\bottomrule
	\end{tabular}}
    \vspace{-2em}
\end{table}

\subsection{Ablation on Different Robot Embodiments
}

\begin{figure}[t]
	\centering
	\includesvg[width=\linewidth]{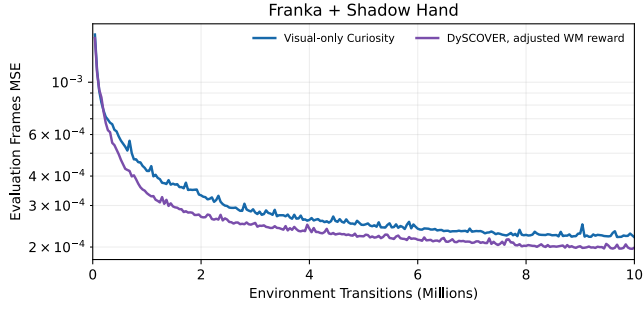}
	\caption{Evaluation frame reconstruction MSE during 10 million training steps for the Franka robot. Lower is better. DySCOVER shows similar performance improvement over the visual-only curiosity baseline.}
	\label{fig:robot-mse}
\end{figure}

We also evaluate the exploration strategies on another robot embodiment to verify that the results remain generic. Figure~\ref{fig:robot-mse} compares visual-only curiosity and DySCOVER on a Franka robot with ``Shadow Hand'' gripper, using the same training settings otherwise. Table ~\ref{tab:reconstruction-small-buffer-robots} shows that at 10M simulation steps, DySCOVER yields a lower evaluation reconstruction error than the visual-only baseline for the Franka embodiment as well, yielding a $9.5\%$ relative improvement. We also performed preliminary testing on a UR10e robot equipped with a different gripper than the one used in the primary experiments, following the same trend: DySCOVER improved visual reconstruction over visual-only curiosity (at $2$M steps, visual-only curiosity achieves $4.30  \times 10^{-4}$, while DySCOVER achieves $4.17 \times 10^{-4}$, giving an improvement of $3.2\%$).

\begin{table}[t]
	\centering
	\caption{Embodiment comparison}
	\label{tab:reconstruction-small-buffer-robots}
	\small
	\resizebox{\columnwidth}{!}{%
		\begin{tabular}{ccccc}
			\toprule
			Enviroment & \makecell{Method} & \makecell{MSE $\downarrow$} & \makecell{PSNR (dB) $\uparrow$} & \makecell{MSE reduction \\ w.r.t. baseline} $\uparrow$ \\
			\midrule
			UR10e & Visual-only curiosity & $3.34\times10^{-4}$ & $34.76$ & --- \\
			UR10e & \textbf{DySCOVER} & $\mathbf{2.83\times10^{-4}}$ & $\mathbf{35.49}$ & $\mathbf{15.3\%}$ \\
			\midrule
			Franka & Visual-only curiosity & $2.20\times10^{-4}$ & $36.57$ & --- \\
			Franka & \textbf{DySCOVER} & $\mathbf{1.99\times10^{-4}}$ & $\mathbf{37.02}$ & $\mathbf{9.5\%}$ \\
			\bottomrule
	\end{tabular}}
    \vspace{-1em}
\end{table}

\subsection{Adaptation to Object-Augmented Scenes}


\noindent\textbf{Object-scene adaptation.}
We examine whether DySCOVER remains useful when the visual distribution changes and becomes richer by adding objects to the scene. The goal is still to improve the agent's vision model, but the agent now has to adapt from handling observations of its body in an empty scene to observations that also include previously unseen objects.

We continue training in the object-augmented scene for an additional million steps. Visual-only baseline and DySCOVER are initialized at their final model checkpoints from the object-free experiments, corresponding to a natural transfer setting from self-body exploration to a richer scene. We also include a visual-only curiosity experiment that is initialized from the same visual model as DySCOVER, as a sanity check, in order to rule out the possibility that its superior performance in the new scene is explained only by a better initial visual representation.

\begin{figure}[t]
	\centering
	\includesvg[width=\columnwidth]{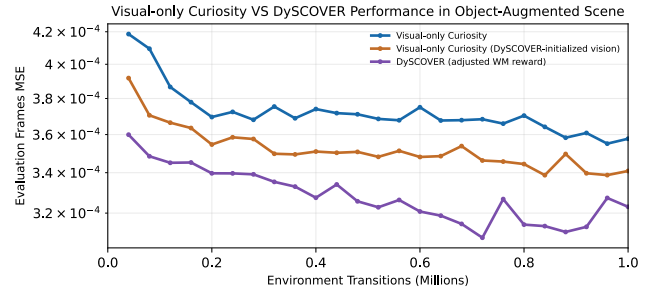}
    \vspace{-1em}
    \caption{Evaluation reconstruction error during object-scene adaptation. DySCOVER reaches lower error than visual-only curiosity, and also remains below the visual-only control initialized from the same visual model.}
	\label{fig:object-scene-mse}
\end{figure}

As Figure~\ref{fig:object-scene-mse} shows, DySCOVER offers a relative MSE reduction of $9.70\%$ with respect to visual-only curiosity. Furthermore, the visual-only curiosity initialized with the same vision model on the object-free experiment remains $5.23\%$ worse than DySCOVER. This suggests that the world model reward helps during adaptation and not only through the initialization of the visual representation.

\begin{figure}[t]
	\centering
	\begin{minipage}[b]{0.32\linewidth}
		\centering
		\footnotesize\strut Ground Truth \\[1mm]
		\includegraphics[width=\linewidth]{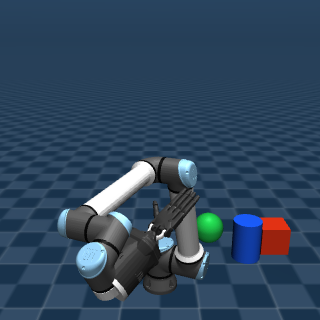}
	\end{minipage}
	\hfill
	\begin{minipage}[b]{0.32\linewidth}
		\centering
		\footnotesize\strut Visual-only Curiosity \\[1mm]
		\includegraphics[width=\linewidth]{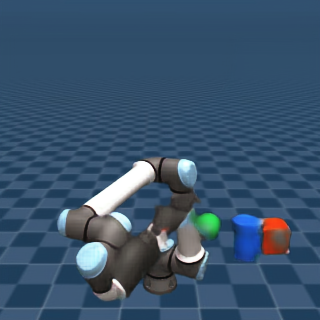}
	\end{minipage}
	\hfill
	\begin{minipage}[b]{0.32\linewidth}
		\centering
		\footnotesize\strut DySCOVER \\[1mm]
		\includegraphics[width=\linewidth]{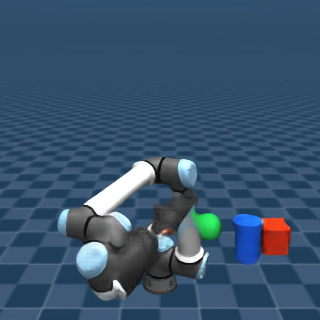}
	\end{minipage}
	
	\vspace{2mm}
	
	\begin{minipage}[b]{0.32\linewidth}
		\centering
		\includegraphics[width=\linewidth]{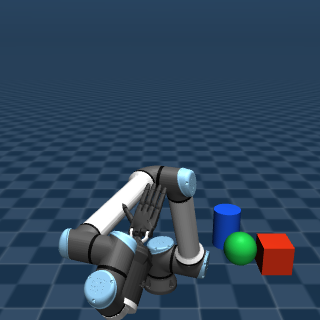}
	\end{minipage}
	\hfill
	\begin{minipage}[b]{0.32\linewidth}
		\centering
		\includegraphics[width=\linewidth]{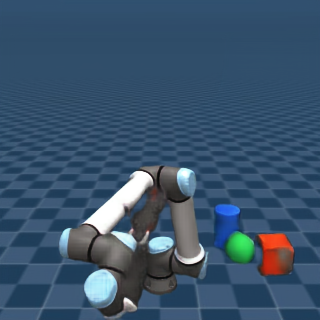}
	\end{minipage}
	\hfill
	\begin{minipage}[b]{0.32\linewidth}
		\centering
		\includegraphics[width=\linewidth]{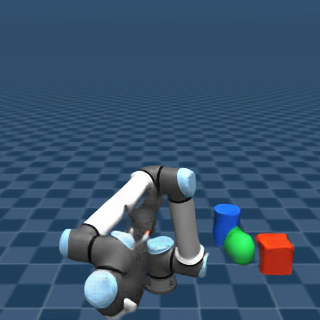}
	\end{minipage}
	
	\caption{Qualitative reconstruction comparison on object-scene frames. DySCOVER performs better in difficult scenes with occlusions.}
	\label{fig:object-scene-qualitative}
\end{figure}

\noindent\textbf{Region-specific reconstruction.}
Moreover, we evaluate on an additional set of frames together with segmentation masks for the robot and the objects, allowing us to isolate the reconstruction error on these regions, rather than letting the background dominate the metric, and therefore determine where the improvement comes from. As shown in Table~\ref{tab:object-scene-reconstruction}, DySCOVER reduces global reconstruction MSE by $5.63\%$ relative to visual-only curiosity. The improvement is greater on the masked robot and object regions: robot-region MSE is reduced by $24.31\%$, and object-region MSE is reduced by $16.99\%$. This is also shown qualitatively in Figure~\ref{fig:object-scene-qualitative}, where DySCOVER better preserves the robot and occluded object regions in two difficult object-scene examples. The object-focused improvement is particularly important because the objects are not part of the original self-body training distribution, suggesting that dynamics-aware curiosity helps the agent collect experience, improving its visual understanding of both itself and newly introduced objects.

\begin{table}[t]
	\centering
    \vspace{-.5em}
	\caption{Object-scene reconstruction performance}
	\label{tab:object-scene-reconstruction}
	\small
	\resizebox{\columnwidth}{!}{%
		\begin{tabular}{lccc}
			\toprule
			Method & Global MSE $\downarrow$ & Robot MSE $\downarrow$ & Object MSE $\downarrow$ \\
			\midrule
			Visual-only curiosity & $3.460{\times}10^{-3}$ & $1.032{\times}10^{-2}$ & $1.256{\times}10^{-2}$ \\
			DySCOVER & $\mathbf{3.265{\times}10^{-3}}$ & $\mathbf{7.809{\times}10^{-3}}$ & $\mathbf{1.043{\times}10^{-2}}$ \\
			\midrule
			\textit{DySCOVER Improvement} & \textit{5.6\%} & \textit{24.3\%} & \textit{17.0\%} \\
			\bottomrule
	\end{tabular}}
    \vspace{-1.1em}
\end{table}

\noindent\textbf{Object-interaction behavior.} We further examine whether these reconstruction gains are accompanied by object-relevant interaction behavior. Table~\ref{tab:object-scene-interactions} shows that DySCOVER's exploration leads to increased object interactions by $18.2\%$, increased maximum object displacement by $47.3\%$, and slightly higher entropy for object-motion directions by $6.3\%$, relative to the visual-only curiosity baseline. These statistics explain the improved object-focused reconstruction seen in Table ~\ref{tab:object-scene-reconstruction}: DySCOVER exposes the visual model to more diverse object-motion events, due to the additional world model curiosity, resulting in experience that is more informative for adapting the visual model to the introduced objects.

All values are averaged over the training cycles.``Objects displaced $>5$ cm'' counts how many times an object's movement exceeded $5$ cm, ``Max object displacement'' is the maximum displacement of any object in a rollout, and ``Motion-direction entropy'' is the normalized entropy of the binned object-motion directions, where higher values show that motion is spread over more directions.

\begin{table}[t]
	\centering
	\caption{Object-interaction average statistics per cycle.}
	\label{tab:object-scene-interactions}
	\small
	\resizebox{\columnwidth}{!}{%
		\begin{tabular}{lccc}
			\toprule
			Metric & Visual-only & DySCOVER & Change \\
			\midrule
			$\#$Objects displaced $>5 \; cm$ & 4.628 & 5.468 & $+18.2\%$ \\
			Max object displacement (cm) & 12.2 & 17.9 & $+47.3\%$ \\
			Motion-direction entropy & 0.796 & 0.846 & $+6.3\%$ \\
			\bottomrule
	\end{tabular}}
    \vspace{-1.5em}
\end{table}

\begin{figure}[t]
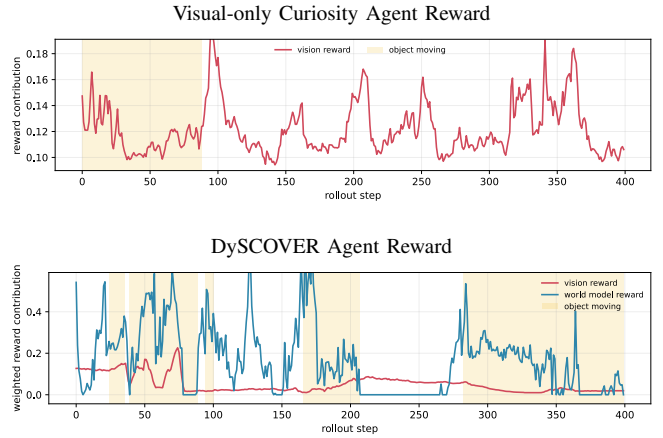

	\centering

	\footnotesize Visual-only Curiosity Agent Reward\\[1mm]
	\includesvg[width=\columnwidth]{images/reward_differences/visual_only/rollout_003_env01_dim00_rewards.svg}

	\vspace{1em}

	\footnotesize DySCOVER Agent Reward\\[1mm]
	\includesvg[width=\columnwidth]{images/reward_differences/attached_rollout_rewards_weighted_shared_actual_weights.svg}

	\caption{Reward decomposition during an example object-scene rollout. The top plot shows the reward per step for a rollout for the visual-only curiosity agent. The bottom plot decomposes the intrinsic reward into the visual-reward term and the WM-reward term; shaded regions indicate intervals where an object is moved by the agent. The WM reward tends to increase during object movement intervals while the reconstruction-based reward does not correlate with them.}
	\label{fig:object-scene-reward-terms}
    \vspace{-1em}
\end{figure}

\noindent\textbf{Reward interpretation.}
Figure~\ref{fig:object-scene-reward-terms} further illustrates why the WM reward can lead to more informative experience. In an example rollout, the visual reward is not necessarily aligned with the intervals where an object is moving, since it measures frame novelty. The WM reward, however, often increases around these object-motion intervals, because it is sensitive to the visual consequences of the agent's actions. This shows that the predictive term helps the policy collect transitions that visual novelty alone may not emphasize.

\subsection{How Dynamics-Aware Reward Affects Policy Behavior}

The previous sections quantitatively show that adding dynamics-aware curiosity improves the final visual model. We now also qualitatively examine how the different reward definitions affect the policy behavior that generates data to be used for model training.

\begin{figure*}[t]
\centering
\includesvg[width=.92\textwidth]{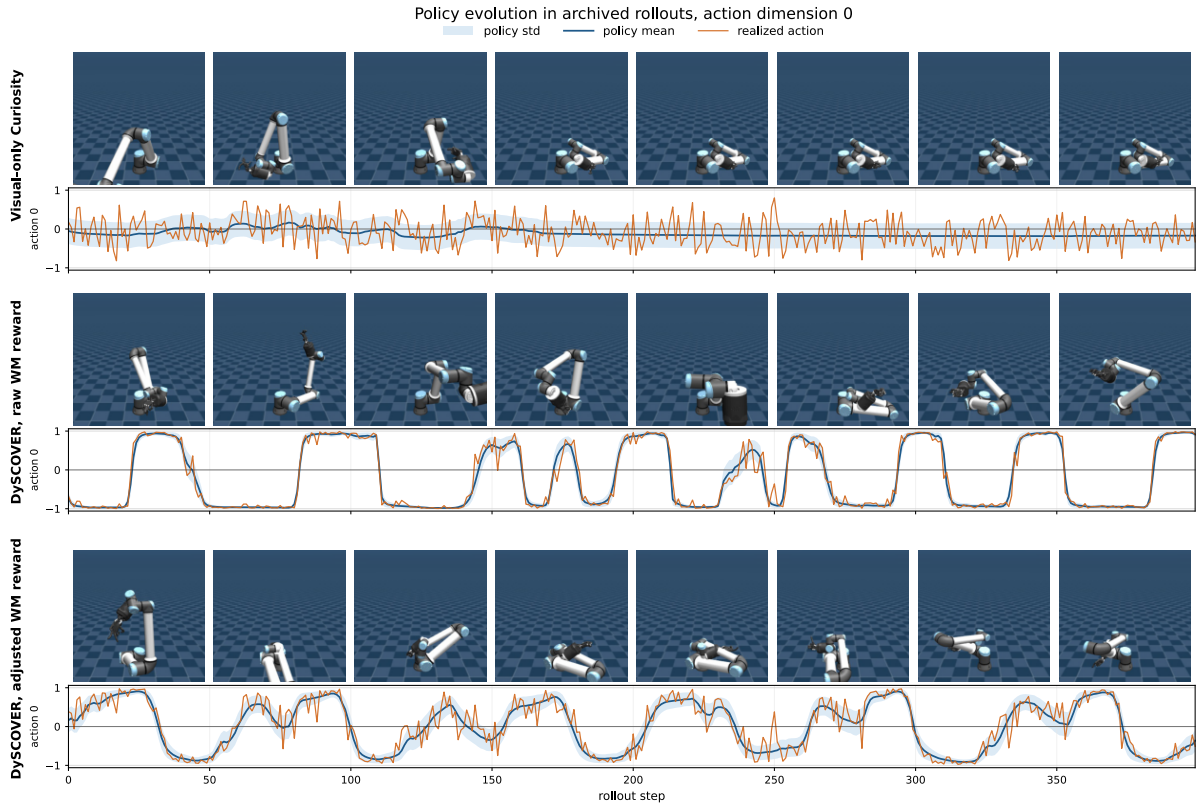}
\vspace{-1.5em}
\caption{Example of policy rollout comparison at 4 million training steps. Each row shows sampled frames from a rollout and policy statistics for an action dimension. The blue curve is the policy mean, the shaded area the policy's standard deviation, and the orange curve is the sampled action. DySCOVER exhibits more complex control than the baseline, producing richer robot configurations and exposing the vision model to more diverse views. Note, DySCOVER with adjusted WM term produces much smoother control, yet complex enough to collect useful frames for the vision model.}
\label{fig:policy-exploration}
\vspace{-1em}
\end{figure*}

Figure~\ref{fig:policy-exploration} compares example rollouts during RL training at the same train step for visual-only curiosity, DySCOVER with the raw WM reward and DySCOVER with the adjusted WM reward. Each row shows sampled observed frames from one rollout and, below them, the policy statistics for an action dimension. The blue curve is the policy mean, the shaded region is the policy's standard deviation, and the orange curve is the sampled action executed during the rollout.

The visual-only curiosity policy produces a comparatively limited control pattern, leading to less diverse robot configurations. This is expected, since the reward depends only on frame-wise reconstruction error and not whether a transition has informative consequences. In contrast, DySCOVER with the raw WM reward produces stronger and more complex action excursions. These motions expose the visual model to richer configurations as seen in the accompanying rollout frames. However, the rollout also shows that raw prediction error can encourage high-magnitude and less smooth behavior: large actions produce large transitions in latent and visual space, which can increase the prediction error and therefore the reward, even when those transitions are not necessarily the most informative for the visual model.

DySCOVER with the adjusted WM reward mitigates this effect. Its policy still generates sufficiently complex motions to result in useful observations, but the control is smoother than in the raw WM reward case. This is expected when using the WM error correction: instead of simply rewarding large prediction errors, it biases the policy towards those prediction errors that are larger than an expected persistent, and possibly irreducible at that moment, error caused by model imperfections.

Overall, the visualization supports the conclusion of the quantitative analysis; dynamics-aware curiosity improves visual learning by enhancing the data-collection policy.

\section{Conclusion and Future Work}

We presented DySCOVER, an active visual-learning loop in which a visual reconstruction curiosity signal is augmented by the prediction error of a learned action-conditioned world model. The world model is used to enhance the reward that the policy uses to collect experience that improves the agent's visual model.

Across multiple experiments, we show that adding the WM reward term improves the reconstruction ability of the vision model, relative to visual-only curiosity. The compact-buffer and cross-robot experiments show that the effect is not limited to the main setting. Furthermore, the object-augmented experiment suggests that the same mechanism can also improve adaptation when the visual distribution changes and becomes more complex. The qualitative analysis of the policies provides a visualization for the way the WM reward term biases the policy towards more complex behaviors that tend to produce more complex and diverse rollouts than visual-only curiosity.

Future work includes the evaluation of the agent on external tasks to investigate whether the visual representation that DySCOVER learns can be used in a complex scenario, such as planning or manipulation. Practical applications that could be investigated are \textit{autonomous exploration of unknown environments}, where a robot must decide how to move in order to enrich its visual representations and possibly build a map of its surroundings, and \textit{object discovery in unstructured environments}, where a robot must learn to detect, distinguish, and represent possibly unseen objects through self-supervised interaction rather than predefined labels or task-specific supervision.

\section*{Acknowledgment}
OpenAI Codex was used to assist with the implementation of portions of the experimental code. All generated code was reviewed and modified by the authors as necessary.

\bibliographystyle{IEEEtran}
\vspace{-.5em}
\bibliography{references}
\end{document}